\documentclass[preprint,12pt]{elsarticle}
\usepackage[pagebackref=true,breaklinks=true,letterpaper=true,colorlinks,bookmarks=false]{hyperref}
\usepackage[utf8]{inputenc}
\usepackage{subfig}
\usepackage{amsthm}
\usepackage{algorithmic}
\usepackage{amssymb}
\usepackage{multirow} 
\usepackage[ruled]{algorithm2e} 
\usepackage{amsmath}
\usepackage{graphicx}
\renewcommand{\vec}[1]{\mathbf{#1}}
\usepackage{array,tabularx}
\usepackage{enumitem}
\newcolumntype{C}[1]{>{\centering\arraybackslash}m{#1}}
\usepackage{epsfig}
\begin{document}

%\title{Structure Learning with Similarity Preserving}
%
%\author{Zhao~Kang, Xiao~Lu, Yiwei~Lu,
%       Chong~Peng,
%Steven C.H. Hoi,~\IEEEmembership{Fellow,~IEEE}, 
%        and~Zenglin~Xu% <-this % stops a space
%\IEEEcompsocitemizethanks{\IEEEcompsocthanksitem Z. Kang, X. Lu, Y. Lu, and Z. Xu are with the School of Computer Science and Engineering, University of Electronic Science and Technology of China, Chengdu, Sichuan, 611731.
%E-mail: \{zkang,~zlxu\}@uestc.edu.cn
%\IEEEcompsocthanksitem C. Peng is with College of Computer Science and Technology, Qingdao University, Qingdao, China, 266071.
%E-mail: {cpeng@qdu.edu.cn}
%\IEEEcompsocthanksitem S. C.H. Hoi is with the School of Information Systems, Singapore Management University, Singapore 188065, Singapore.
%E-mail: {chhoi@smu.edu.sg}
%%\IEEEcompsocthanksitem F. Nie is with the School of Computer Science, Northwestern Polytechnical University, Xi’an 710072, China, and also with the Center for OPTical IMagery Analysis and Learning, Northwestern Polytechnical
%%University, Xi’an 710072, China.
%%E-mail: { feipingnie@gmail.com}
%% Department of Computer Science, University of Manitoba, 66 Chancellors Cir, Winnipeg, MB R3T 2N2, Canada
%}\\
\begin{frontmatter}
\title{Structure Learning with Similarity Preserving}
\author[label1]{Zhao Kang}
\author[label1]{Xiao~Lu}
\author[label1]{Yiwei~Lu}
\author[label2]{ Chong~Peng}
%\author[label3]{Steven C.H. Hoi}
\author[label1]{Zenglin Xu}%\corref{mycorrespondingauthor}}
\address[label1]{School of Computer Science and Engineering,\\
University of Electronic Science and Technology of China, Sichuan, 611731, China.}
\address[label2]{College of Computer Science and Technology, Qingdao University, Qingdao, 266071, China.}
%\address[label3]{School of Information Systems, Singapore Management University, Singapore 188065, Singapore.}
%\address[label4]{Institute of High Performance Computing, A*STAR,Singapore}
%\address[label5]{College of Computer Science, Sichuan University, China}




%\maketitle

\begin{abstract}
Leveraging on the underlying low-dimensional structure of data, low-rank and sparse modeling approaches have achieved great success in a wide range of applications. However, in many applications the data can display structures beyond simply being low-rank or sparse. Fully extracting and exploiting hidden structure information in the data is always desirable and favorable. To reveal more underlying effective manifold structure, in this paper, we explicitly model the data relation. Specifically, we propose a structure learning framework that retains the pairwise similarities between the data points. Rather than just trying to reconstruct the original data based on self-expression, we also manage to reconstruct the kernel matrix, which functions as similarity preserving. Consequently, this technique is particularly suitable for the class of learning problems that are sensitive to sample similarity, e.g., clustering and semisupervised classification. To take advantage of representation power of deep neural network, a deep auto-encoder architecture is further designed to implement our model. Extensive experiments on benchmark data sets demonstrate that our proposed framework can consistently and significantly improve performance on both evaluation tasks. We conclude that the quality of structure learning can be enhanced if similarity information is incorporated.
\end{abstract}

%\begin{IEEEkeywords}
%Similarity preserving, clustering, semisupervised classification, similarity measure, deep auto-encoder.
%\end{IEEEkeywords}


\end{frontmatter}

% \fi
%\IEEEpeerreviewmaketitle

\section{Introduction}
With the advancements in information technology, high-dimensional data become very common for representing the data. However, it is difficult to deal with high-dimensional data due to challenges such as the curse of dimensionality, storage and computation costs. Fortunately, in practice data are not unstructured. For example, their samples usually lie around low-dimensional manifolds and have high correlation among them \cite{liu2013robust,kang2015pca}. This phenomenon is validated by the widely used Principal Component Analysis (PCA) where the number of principal components is much smaller than the data dimension. Such a phenomenon is also evidenced in nonlinear manifold learning \cite{lee2007nonlinear}. Since dimension is closely related to the rank of matrix, low-rank characteristic has been shown to be very effective in studying the low-dimensional structures in data \cite{peng2015subspace,zhang2016joint}.

%\begin{equation*}
%\min_W \quad  Tr(WXLX^TW^T) \quad s.t. \quad WXDX^TW^T=I
%\end{equation*}

Another motivation of utilizing rank in data and signal analysis is due to the tremendous success of sparse representation \cite{Elad2010} and compressed sensing \cite{donoho2006compressed}, which are mainly applied to deal with first order data, such as voices and feature vectors. As an extension to the sparsity of order one data, low rankness is a measure for the sparsity of second order data, such as images \cite{Lin2017}. Low-rank models can effectively capture the correlation among rows and columns of a matrix as shown in robust PCA \cite{candes2011robust}, matrix completion \cite{candes2009exact,kang2016AAAI}, and so on. Recently, low-rank and sparse models have shown their effectiveness in processing high-dimensional data by effectively extracting rich low-dimensional structures in data, despite gross corruption and outliers. Unlike traditional manifold learning, this approach often enjoys good theoretical guarantees. 

When data resides near multiple subspaces, a coefficient matrix $Z$ is introduced to enforce correlation among samples. Two typical models are low-rank representation (LRR) \cite{liu2013robust} and sparse subspace clustering (SSC) \cite{elhamifar2013sparse}. Both LRR and SSC aim to find a coefficient matrix $Z$ by trying to reconstruct
each data point as a linear combination of all the other data
points, which is called self-expression property. $Z$ is assumed to be
low-rank in LRR and sparse in SSC. In the literature, $Z$ is also called the similarity matrix since it measures the similarity between samples \cite{chen2015similarity}. LRR and SSC have achieved impressive performance in face clustering, motion segmentation, etc. In these applications, they first learn a similarity matrix $Z$ from the data by minimizing the reconstruction error. After that, they implement spectral clustering by treating $Z$ as similarity graph matrix \cite{peng2017constructing}. Self-expression idea inspires a lot of work along this line. Whenever similarity among samples/features is needed, it can be used. For instance, in recommender system, we can use it to calculate the similarity among users and items \cite{ning2011slim}; in semisupervised classification, we can utilize it to obtain the similarity graph \cite{li2015learning}; in multiview learning, we can use it to characterize the connection between different views \cite{taoensemble2017}. 

More importantly, there are a variety of benefits to obtain similarity matrix through self-expression. First, by this means, the most informative “neighbors” for each data point are automatically chosen and the global structure information hidden in the data is explored \cite{nie2014clustering}. This will avoid many drawbacks in widely used $k$-nearest-neighborhood and $\epsilon$-nearest-neighborhood graph construction methods, such as determination of neighbor number $k$ or radius $\epsilon$. Second, it is independent of similarity metrics, such as Cosine, Euclidean distance, Gaussian function, which are often data-dependent and sensitive to noise and outliers \cite{kang2017twin,tang2019unsupervised}. Third, this automatic similarity learning from data can tackle data with structures at different scales of size and density \cite{huang2015new}. Therefore, low-rank and sparse modeling based similarity learning can not only unveil low-dimensional structure, but also be robust to uncertainties of real-world data. It dramatically reduces the potential chances that might heavily influence the subsequent tasks \cite{zelnik2004self}. 
 
Nevertheless, the data in various real applications is usually very complicated and can display structures beyond simply being low-rank or sparse \cite{haeffele2014structured}. Hence, it is essential to learn the representation that can well embed the rich structure information in the original data. Existing methods usually employ some simple models, which is generally less effective and hard to capture such rich structural information that exists in real world data. To combat this issue, in this paper we demonstrate that it is beneficial to preserve similarity information between samples when we perform structure learning and design a novel term for this task. This new term measures the inconsistency between two kernel matrices, one for raw data and another for reconstructed data, such that the reconstructed data well preserves rich structural information from the raw data. The advantage of this approach is demonstrated in three important problems: shallow clustering, semi-supervised classification, and deep clustering. 

Compared with existing work in the literature, the main contributions of this paper are as follows:%To the best of our knowledge, this is the first work that tries to preserve similarity information in low-dimensional structure learning. 
\begin{itemize}
\item{Different from current low-dimensional structure learning methods, we explicitly model the data relation by preserving the pairwise similarity of the original data with a novel term. Our approach reduces the inconsistency between the structural information of raw and reconstructed data, which leads to enhanced performance.  }
\item{Our proposed structure learning framework is also applied to deep auto-encoder. This helps to achieve a more informative and discriminative latent representation.}
\item{The effectiveness of the proposed approach is evaluated on both shallow and deep models with tasks from image clustering, document clustering, face recognition, digit/letter recognition, to visual object recognition. Comprehensive experiments demonstrate the superiority of our technique over other state-of-the-art methods. }
\item{Our method can serve as a fundamental framework, which can be readily applied to other self-expression methods. Moreover, beyond clustering and classification applications, the proposed framework can be efficiently generalized to a variety of other learning tasks.  }
\end{itemize}

The rest of the paper is organized as follows. Section \ref{background} gives a brief review about two popular algorithms. Section \ref{proposed} introduces the proposed technique and discusses its typical applications to spectral clustering and semi-supervised classification tasks. After that, we present a deep neural network implementation of our technique in Section \ref{deep}. Clustering and semi-supervised classification experimental results and analysis are presented in Section \ref{shallowexp} and Section \ref{semiexp}, respectively. Section \ref{deepexp} validates our proposed deep clustering model. Finally, Section \ref{conclusion} draws conclusions.

\textbf{Notations.} Given a data matrix $X\in\mathcal{R}^{m \times n}$ with $m$ features and $n$ samples, we denote its $(i,j)$-th element and $i$-th column as $x_{ij}$ and $x_i$, respectively. The $\ell_2$-norm of vector \textbf{$x$} is represented as \textbf{$\|x\|=\sqrt{x^\top\cdot x}$}, where $\top$ is transpose operator. The $\ell_1$-norm of $X$ is denoted by $\|X\|_1=\sum_{ij}|x_{ij}|$.  The squared Frobenius norm is defined as $\|X\|_F^2=\sum_{ij}x_{ij}^2$. The definition of $X$'s nuclear norm is $\|X\|_*=\sum_i\sigma_i$, where $\sigma_i$ is the $i$-th singular value of $X$. $I$ represents the identity matrix with proper size and $\vec{1}$ denotes a column vector with proper length where all elements are ones. $Z\geq 0$ means all the elements of $Z$ are nonnegative. Inner product is denoted by $<x_i, x_j>=x_i^\top\cdot x_j$. Trace operator is denoted by $Tr(\cdot)$.
\section{Related Work}
\label{background}
In this paper, we focus on the learning of new representation that characterizes the relationship between samples, namely, the pairwise similarity information. It is well-known that similarity measure is a fundamental and crucial problem in machine learning, pattern recognition, computer vision, data mining and so on \cite{zuo2018guest,liu2015absent}. A number of traditional approaches are often utilized in practice for convenience. As aforementioned, they often suffer from different kinds of drawbacks. Adaptive neighbors approach can learn a similarity matrix from data, but it can only capture the local structure information and thus the performance might have deteriorated in clustering \cite{nie2014clustering}. 

% In this section, we give a brief overview about two popular similarity learning techniques which have been developed recently. 
%\subsection{Adaptive Neighbors Approach}  % 
%In a similar spirit of locality preserving projection (LPP) \cite{niyogi2004locality} method, adaptive neighbors approach assigns a probability $z_{ij}$ for $x_j$ as the neighborhood of $x_i$. In a sense, $z_{ij}$ characterizes the similarity between $x_i$ and $x_j$ \cite{nie2014clustering}. The smaller distance should be $\|x_i-x_j\|^2$, the greater the probability $z_{ij}$ is, and the more similar $x_i$ and $x_j$ are. By optimizing the following problem:
%\begin{equation}
%\begin{split}
%&\min_{z_i} \sum_{j=1}^n (\| x_i-x_j\|^2z_{ij}+\alpha z_{ij}^2), \\
%&\quad\hspace{.1cm} s.t. \hspace{.1cm} z_i^\top\vec{1}=1, \hspace{.1cm}  0\leq z_{ij}\leq 1,
%\end{split}
%\label{local}
%\end{equation}
%where $\alpha$ is a trade-off parameter, one can learn $Z$ adaptively from the data. Compared to LPP, where neighborhood relationship is predetermined, the structure is automatically learned from the data in this approach. This idea has been applied in a number of problems recently. Nonnegative matrix factorization \cite{dacheng2017}, feature selection \cite{du2015unsupervised}, multiview graph learning \cite{nie2017multi}, just to name a few. A limitation of this method is that it can only capture the local structure information and thus the performance might have deteriorated. 

%\subsection{Self-expression Approach} %he2011nonnegative, yang2014semi,
Self-expression, another strategy, has become increasingly popular in recent years \cite{yang2014data}. The basic idea is to encode each datum as a weighted combination of other samples, i.e., its direct neighbors and reachable indirect neighbors. Similar to locally linear embedding (LLE) \cite{roweis2000nonlinear}, if $x_i$ and $x_j$ are similar, weight coefficient $z_{ij}$ should be big. From this point of view, $Z$ also behaves like a similarity matrix. For convenience, we denote the reconstructed data as $\hat{X}$, where $\hat{X}=XZ$. The discrepancy between the original data $X$ and the reconstructed data $\hat{X}$ is minimized by solving the following problem:
\begin{equation}
\begin{split}
&\min_{Z}\frac{1}{2} \|X-XZ\|_F^2+\beta \rho(Z),\hspace{.1cm}\\
%&\quad\quad s.t. \hspace{.1cm} Z\geq 0,
&\quad\quad s.t. \hspace{.1cm} diag(Z)= 0,\end{split}
\label{global}
\end{equation} 
where $\rho(Z)$ is a regularizer on $Z$, $\beta>0$ is used to balance the effects of the two terms. Thus, we can seek either a sparse representation or a low-rank representation of the data by adopting the $\ell_1$ norm and nuclear norm of $Z$, respectively. Since this approach can capture the global structure information hidden in the data, it has drawn significant attention and achieved impressive performance in a number of applications, including face recognition \cite{zhang2011sparse}, subspace clustering \cite{yao2019multiview,elhamifar2013sparse,feng2014robust}, semisupervised learning \cite{zhuang2012non}, dimension reduction \cite{lu2016low}, and vision learning \cite{li2017self}. To consider nonlinear or manifold structure information of data, some kernel-based methods \cite{xiao2016robust,patel2014kernel} and manifold learning methods \cite{zhuang2016locality,liu2014enhancing} have been developed. However, these manifold-based methods depend on labels or graph Laplacian, which are often not available.  

Recently, Kang et al. propose a twin learning for similarity and clustering (TLSC) \cite{kang2017twin} method. TLSC performs similarity learning and clustering in a unified framework. In particular, the similarity matrix is learned via self-expression in kernel space. Consequently, it shows impressive performance in clustering task.

However, all existing self-expression based methods just try to reconstruct the original data such that some valuable information is largely ignored. In practice, the low-dimensional manifold structure of real data is often very complicated and presents complex structure apart from low-rank or sparse \cite{haeffele2014structured}. Exploiting data relations has been proved to be a promising means to discover the underlying structure in a number of techniques \cite{tenenbaum2000global,roweis2000nonlinear}. For instance, ISOMAP \cite{tenenbaum2000global} retains the geodesic distance between pairwise data in the low-dimensional space. LLE \cite{roweis2000nonlinear} learns a low-dimensional manifold by preserving the linear relation, i.e., each data point is a linear combination of its neighbors. To seek a low-dimensional manifold, Laplacian Eigenmaps \cite{belkin2002laplacian} minimizes the weighted pairwise distance in the projected space, where weight characterizes the pairwise relation in the original space. 

In this paper, we demonstrate how to integrate similarity information into the construction of new representation of data, resulting in a significant improvement on two fundamental tasks, i.e., clustering and semi-supervised classification. More importantly, the proposed idea can be readily applied to other self-expression methods such as smooth representation \cite{hu2014smooth}, least squared representation \cite{lu2012robust}, and many applications, e.g., Occlusion Removing \cite{qian2015robust}, Saliency Detection \cite{lang2012saliency}, Image Segmentation \cite{cheng2011multi}.

\section{Proposed Formulation}
\label{proposed}
To make our framework more general, we build our model in kernel space. Eq.(\ref{global}) can be easily extended to kernel representation through mapping $\phi$. By utilizing kernel trick $k(x,y)=\phi(x)^\top\phi(y)$, we have
 \begin{equation}
\begin{split}
&\min_{Z} \frac{1}{2}\|\phi(X)-\phi(X)Z\|_F^2+\beta \rho(Z),\\
\Longleftrightarrow&\min_{Z} \frac{1}{2}Tr(\phi(X)^\top\phi(X)-\phi(X)^\top\phi(X) Z\\
&-Z^\top\phi(X)^\top \phi(X)+Z^T\phi(X)^\top\phi(X) Z)+\beta \rho(Z),\\
\Longleftrightarrow
&\min_{Z}\frac{1}{2} Tr(K-2KZ+Z^{\top}KZ)+\beta \rho(Z),\hspace{.1cm}\\
&\quad\quad s.t. \hspace{.1cm} diag(Z)= 0.
%&\quad\quad s.t. \hspace{.1cm} Z\geq 0,
\end{split}
\label{kernel}
\end{equation} 
By solving this problem, we can learn the nonlinear relations among $X$. Note that (\ref{kernel}) becomes (\ref{global}) if a linear kernel is adopted. 

In this paper, we aim to preserve the similarity information of the original data. To this end, we make use of the widely used inner product. Specifically, we try to minimize the inconsistency between two inner products: one for the raw data and another for reconstructed data $XZ$. To make our model more general, we build it in a transformed space. In other words, we have
\begin{equation}
\label{pointsim}
\min_Z \|\phi(X)^\top\cdot\phi(X)-(\phi(X)Z)^\top\cdot(\phi(X)Z)\|_F^2.
\end{equation}
 (\ref{pointsim}) can be simplified as 
\begin{equation}
\min_Z \|K-Z^\top KZ\|_F^2.
\label{preserve}
\end{equation}
%If two points $i$ and $j$ are from distinct clusters, their similarity $K_{ij} = 0$. Idealy, $Z_{ij}$ should be zero. Therefore, (\ref{preserve}) is generally true. $X^{\top}X$ is nothing but the linear kernel of the original data, so we can substitute it with general kernel matrix $K$, which leads to  $\|K-Z^\top KZ\|_F^2$. 
Comparing Eq. (\ref{preserve}) to (\ref{kernel}), we can see that Eq. (\ref{preserve}) involves higher order of $Z$. Thus, our designed Eq. (\ref{preserve}) captures high order information of original data. Although we claim that our method seeks to preserve similarity information, it also includes dissimilarity preserving effect, so it can preserve the relations between samples in general. 
 Combining (\ref{preserve}) with (\ref{kernel}), we obtain our \textbf{S}tructure \textbf{L}earning with \textbf{S}imilarity \textbf{P}reserving (SLSP) framework:
 \begin{equation}
\begin{split}
&\min_{Z}\frac{1}{2}Tr(\!K\!-\!2KZ\!+\!Z^{\top\!}KZ\!)\!+\!\alpha\|\!K\!-\!Z^\top KZ\!\|_F^2\!+\!\beta\rho(Z),\\
&\quad s.t.  \quad diag(Z)= 0.
%&\quad s.t.  \quad Z\geq 0.
\end{split}
\label{obj}
\end{equation} 
Through solving this problem, we can obtain either a low-rank or sparse matrix $Z$, which carries rich structure information of the original data. Besides this, SLSP enjoys several other nice properties:\\
(1) Our framework can not only capture global structure information but also preserve the original pairwise similarities between the data points in the original data in the embedding space. If a linear kernel function is adopted in (\ref{obj}), our framework can recover linear structure information hidden in the data. \\%We believe that this is the first study that considers similarity preserving effects in low-dimensional structure learning.
(2) Our proposed technique is particularly suitable to problems that are sensitive to sample similarity, such as clustering \cite{elhamifar2013sparse}, classification \cite{zhuang2012non}, users/items similarity in recommender systems \cite{ning2011slim}, patient/drug similarity in healthcare informatics \cite{zhu2016measuring}. We believe that our framework can effectively model and extract rich low-dimensional structures in high-dimensional data such as images, documents, and videos. \\
(3) The input is kernel matrix. This is an appealing property, as not all types of real-world data can be represented in numerical feature vectors form. For example, we often find clusters of proteins based on their structures and group users in social media according to their friendship relations.\\
(4) Generic similarity rather than inner product can also be used to construct (\ref{preserve}) given that the resulting optimization problem is still solvable. It means that similarity measures that reflect domain knowledge such as \cite{popescu2006fuzzy} can be incorporated in SLSP directly. Even dissimilarity measures can be included in this algorithm. This flexibility extends the range of applicability of SLSP.
\subsection{Optimization}
Although the SLSP problem can be solved in several different  ways, we describe an alternating direction method of multipliers (ADMM) [41] based approach, which is easy to understand. Since the objective function in (\ref{obj}) is a fourth-order function of $Z$, ADMM can lower its order by introducing auxiliary variables.

 First, we rewrite (\ref{obj}) in the following equivalent form by introducing three new variables:    
 \begin{equation}
\begin{split}
&\min_{Z}\frac{1}{2}Tr(\!K\!-\!2KJ\!+\!J^{\top\!}KJ\!)\!+\!\alpha\|\!K\!-\!W^\top KH\!\|_F^2\!+\!\beta\rho(Z),\hspace{.1cm}\\
&\quad \quad s.t. \hspace{.2cm} diag(Z)= 0, \hspace{.1cm}  J=Z,  \hspace{.1cm} W=Z, \hspace{.1cm}  H=Z.
\end{split}
\label{eqlobj}
\end{equation} 
Then its augmented Lagrangian function can be written as:
 \begin{equation}
\begin{split}
&\mathcal{L}(J,W,H,Z,Y_1,Y_2,Y_3)=\\
&\frac{1}{2}Tr(\!K\!-\!2KJ\!+\!J^{\top\!}KJ\!)\!+\!\alpha\|\!K\!-\!W^\top KH\!\|_F^2\!+\!\beta\rho(Z)\!+\\
&\frac{\mu}{2}\left(\|\!J\!-\!Z\!+\!\frac{Y_1}{\mu}\|_F^2\!+\!\|\!W\!-\!Z\!+\!\frac{Y_2}{\mu}\|_F^2\!+\!\|\!H\!-\!Z\!+\!\frac{Y_3}{\mu}\|_F^2\!\right)
%&\quad \quad s.t. \hspace{.2cm} Z\geq 0, \hspace{.1cm}  J=Z,  \hspace{.1cm} W=Z, \hspace{.1cm}  H=Z.
\end{split}
\label{lag}
\end{equation} 
where $\mu>0$ is a penalty parameter and $Y_1$, $Y_2$, $Y_3$ are Lagrangian multipliers. We can update those variables alternatively, one at each step, while keeping the others fixed. Then, it yields the following updating rules.

Updating $J$: By removing the irrelevant terms, we arrive at:
  \begin{equation}
%\begin{split}
\min_{J}\frac{1}{2}Tr(-\!2KJ\!+\!J^{\top\!}KJ\!)\!+\frac{\mu}{2}\|\!J\!-\!Z\!+\!\frac{Y_1}{\mu}\|_F^2
%\end{split}
\end{equation} 
It can be seen that it is a strongly convex quadratic function and can be solved by setting its first derivative to zero, so
\begin{equation}
J=(K+\mu I)^{-1}(K+\mu Z-Y_1).
\label{updateJ}
\end{equation} 

Updating $W$: For $W$, we are to solve:
  \begin{equation}
%\begin{split}.
\min_{W}\alpha\|\!K\!-\!W^\top KH\!\|_F^2+\frac{\mu}{2}\|\!W\!-\!Z\!+\!\frac{Y_2}{\mu}\|_F^2.
%\end{split}
\end{equation} 
By setting its first derivative to zero, we obtain
\begin{equation}
W=(2\alpha KHH^\top K^\top+\mu I)^{-1}(2\alpha KH K^\top+\mu Z-Y_2).
\label{updateW}
\end{equation} 

Updating $H$: We fix other variables except $H$, the objective function becomes:
  \begin{equation}
%\begin{split}
\min_{H}\alpha\|\!K\!-\!W^\top KH\!\|_F^2+\frac{\mu}{2}\|\!H\!-\!Z\!+\!\frac{Y_3}{\mu}\|_F^2.
%\end{split}
\end{equation} 
Similar to $W$, it yields
\begin{equation}
H=(2\alpha K^\top WW^\top K+\mu I)^{-1}(2\alpha  K^\top WK+\mu Z-Y_3).
\label{updateH}
\end{equation}

Updating $Z$: For $Z$, the subproblem is:
  \begin{equation}
\min_{Z}\beta\rho(Z)+\frac{3\mu}{2}\|Z-D\|_F^2,
\end{equation} 
where $D=\frac{J+W+H+\frac{Y_1+Y_2+Y_3}{\mu}}{3}$.
Depending on regularization strategy, we have different closed-form solutions for $Z$. Let's write the singular value decomposition (SVD) of $D$ as $Udiag(\sigma)V^\top$. Then, for low-rank representation, i.e., $\rho(Z)=\|Z\|_*$, we have \cite{cai2010singular},
 \begin{equation}
 Z=U diag(\textit{max}\{\sigma-\frac{\beta}{3\mu},0\}) V^T.
\label{lowrank}
\end{equation}
To obtain a sparse representation, i.e., $\rho(Z)=\|Z\|_1$, we can update $Z$ element-wisely as \cite{beck2009} :
\begin{equation}
 Z_{ij}=max\{|D_{ij}|-\frac{\beta}{3\mu},0\}\cdot \textit{sign}(D_{ij}).
\label{sparse}
\end{equation}
For clarity, the complete algorithm to solve problem (\ref{eqlobj}) is summarized in Algorithm 1. We stop the algorithm if the maximum iteration number 300 is reached or the relative change of $Z$ is less than $10^{-5}$. 
\begin{algorithm}%[!tb]
% \small  %\cite{cai2010singular}  \cite{beck2009} 
\caption{The algorithm of SLSP}
\label{alg2}
 {\bfseries Input:} Kernel matrix $K$, parameters  $\alpha>0$, $\beta>0$, $\mu>0$.\\
{\bfseries Initialize:} Random matrix $H$ and $Z$, $Y_1=Y_2=Y_3=0$.\\
 {\bfseries REPEAT}
\begin{algorithmic}[1]
\STATE Calculate $J$ by (\ref{updateJ}).
 \STATE Update $W$ according to (\ref{updateW}).% correspond to the $c$ smallest eigenvalues.
\STATE Calculate $H$ using (\ref{updateH}) 
\STATE Calculate $Z$ using (\ref{lowrank}) or (\ref{sparse}).
\STATE Update Lagrange multipliers $Y_1$, $Y_2$ and $Y_3$ as
\begin{eqnarray*}
Y_1=Y_1+\mu(J-Z),\\
Y_2=Y_2+\mu(W-Z),\\
Y_3=Y_3+\mu(H-Z).
\end{eqnarray*}
\end{algorithmic}
\textbf{ UNTIL} {stopping criterion is met.}
\end{algorithm}
\subsection{Complexity Analysis}
 First, the construction of kernel matrix costs $\mathcal{O}(n^2)$. The computational cost of Algorithm 1 is mainly determined by updating the variables $J$, $W$, and $H$. All of them involve matrix inversion and multiplication of matrices, whose complexity is $\mathcal{O}(n^3)$. For large scale data sets, we might alleviate this by resorting to some approximation techniques or tricks, e.g., Woodbury matrix identity. In addition, depending on the choice of regularizer, we have different complexity for $Z$. For low-rank representation, it requires an SVD for every iteration and its complexity is $\mathcal{O}(rn^2)$ if we employ partial SVD ($r$ is lowest rank we can find), which can be achieved by package like PROPACK. The complexity of obtaining a sparse solution $Z$ is $\mathcal{O}(n^2)$. The updating of $Y_1$, $Y_2$, and $Y_3$ cost $\mathcal{O}(n^2)$.
\subsection{Application of Similarity Matrix $Z$}
One typical application of $Z$ is spectral clustering which builds the graph Laplacian $L$ based on pairwise similarities between data points. pecific, $L=D-Z$, where $D$ is a diagonal matrix with $i$-th element as $\sum_j z_{ij}$. Spectral clustering solves the following problem:
\begin{equation}
\min_F Tr(F^\top LF), \quad F^\top F=I,
\label{sc}
\end{equation} 
where $F\in \mathcal{R}^{n\times c}$ is the cluster indicator matrix.

Another classical task that make use of $Z$ is semi-supervised classification. In the past decade, graph-based semi-supervised learning (GSSL) has attracted numerous attentions due to its elegant formulation and low computation complexity \cite{cheng2009sparsity}. Similarity graph construction is one of the two fundamental components in GSSL, which is critical to the quality of classification. Nevertheless, with respect to label inference, graph construction has attracted much less attention until recent years \cite{Berton2015}. %In this subsection, we demonstrate that our framework can best capture the essential data structure. 

After we obtain $L$, we can adopt the popular local and global consistency (LGC) as the classification framework \cite{zhou2004learning}. LGC finds a classification function $F\in \mathcal{R}^{n\times c}$ by solving the following problem:
\begin{equation}
\min_F Tr\{F^\top LF+\gamma(F-Y)^\top(F-Y)\},
\label{ssgl}
\end{equation} 
where $c$ is the class number, $Y\in \mathcal{R}^{n\times c}$ is the label matrix, in which $y_{ij}=1$ iff the $i$-th sample belongs to the $j$-th class, and $y_{ij}=0$ otherwise. 

\section{Extension to Deep Model}
\label{deep}
%The proposed objective function in Eq. (\ref{obj}) employs the kernel method to explore the nonlinearity of data. Recently, deep neural network (DNN), another approach to dealing with nonlinear data, has achieved great success in various areas \cite{goodfellow2016deep}. The main advantage of DNN lies in its ability to learn meaningful features from data, which could induce the``intrinsic data structure". Therefore, we can implement our SLSP framework by utilizing a deep neural network. As we show later, the proposed similarity preserving regularizer also enhance the performance in this setting.
The proposed objective function in Eq. (\ref{obj}) can discover the structure in the input space. However, it has less representation powers of data. On the other hand, deep auto-encoder \cite{hinton2006reducing} and its variants \cite{vincent2010stacked,masci2011stacked} can learn structure of data in the nonlinear feature space. However, it ignores the geometry of data in learning data representations. It is a key challenge to learn useful representations for a specific task \cite{bengio2013representation}. In this paper, we propose the idea of similarity preserving for structure learning. Therefore, it is alluring to get the best of both worlds by implementing our SLSP framework within auto-encoder. As we show later, the proposed similarity preserving regularizer indeed enhance the performance of auto-encoder.
\begin{figure*}[!htbp]
\centering
\includegraphics[width=.9\textwidth]{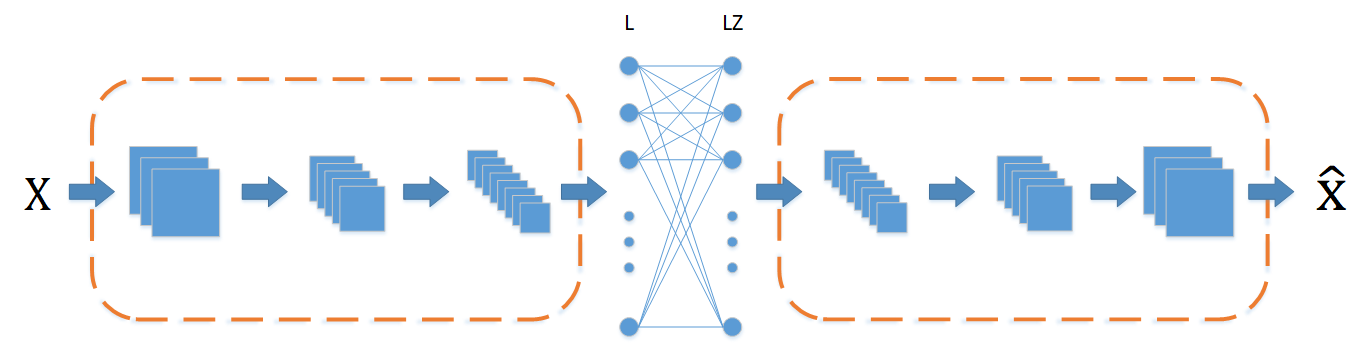}
\caption{The framework of Deep SLSP. Input data $X$ are mapped onto the latent space by the encoder, passed through a fully connected layer to represent $L$ by itself, and finally reconstructed by the decoder.}
\label{net}
\end{figure*}
\subsection{Model Formulation}
Implementing Eq. (\ref{obj}) in auto-encoder, we first need to express $Z$. Recently, Ji et al. \cite{ji2017deep} proposed a deep subspace clustering model with the capability of similarity learning. Inspired by it, we introduce a self-expression layer into the deep auto-encoder architecture. Without bias and activation function, this fully connected layer encodes the notion of self-expression. In other words, this weights of this layer are the matrix $Z$. In addition, kernel mapping is no longer needed since we transform the input data with a neural network. Then, the architecture to implement our model can be depicted as Figure \ref{net}. As we can see, input data $X$ is first transformed into a latent representation $L$, self-expressed by a fully-connected layer, and again mapped onto the original space. 

Let $\hat{X}$ denote the recovered data by decoder. We take each data point $\{l_i\}_{i=1,\cdots,n}$ as a node in the network. Let the network parameters $\Theta$ consist of encoder parameters $\Theta_e$, self-expression layer parameters $Z$, and decoder parameters $\Theta_d$. Then, $\hat{X}$ is a function of $\{\Theta_e, Z, \Theta_d\}$ and $L$ is a function of $\Theta_e$. Eventually, we reach our loss function for Deep SLSP (DSLSP) as: 
 \begin{equation}
\begin{split}
&L(\Theta)=\frac{1}{2}\|X-\hat{X}\|_F^2+\lambda_1 \|L-LZ\|_F^2+\lambda_2 \rho(Z)+\\
&\lambda_3\|L^\top L-Z^\top L^\top LZ\|_F^2\, \quad s.t.  \quad diag(Z)= 0.
%&\quad s.t.  \quad Z\geq 0.
\end{split}
\label{deepobj}
\end{equation} 
The first term denotes the traditional reconstruction loss which guarantees the recovering performance, so that the latent representation will retain the original information as much as possible. With the reconstruction performance guaranteed, the latent representation $L$ can be treated as a good representation of the input data $X$. The second term is the self-expression as in Eq. (\ref{global}). The fourth term is the key component which functions as similarity preserving. For simplicity, it is implemented by dot product. This is also motivated by the fact that our input data points have experienced a series of highly non-linear transformations produced by the encoder. 

%In the traditional autoencoders, $x_i$ is only used to reconstruct itself. In Eq. (\ref{deepobj}), Then we can assume that the data points in this latent space lie in a union of linear subspaces and are well-suited to perform structure learning.

\section{Shallow Clustering Experiment}
\label{shallowexp}
%To demonstrate the effectiveness of the proposed framework, we conduct benchmark experiments in two scenarios, namely, clustering on images/documents and semisupervised classification on recognition tasks.
In this section, we conduct clustering experiments on images and documents with shallow models.% and semisupervised classification on recognition tasks.
%\section{Clustering}
\subsection{Data}
\begin{table}[!htbp]
\centering
\caption{Description of the data sets}
\label{data}
\renewcommand{\arraystretch}{1.3}
\begin{tabular}{l|c|c|c}
\hline
&\textrm{\# instances}&\textrm{\# features}&\textrm{\# classes}\\\hline
\textrm{YALE}&165&1024&15\\\hline
\textrm{JAFFE}&213&676&10\\\hline
\textrm{ORL}&400&1024&40\\\hline
%\textrm{AR}&840&768&120\\\hline
\textrm{COIL20}&1440&1024&20\\\hline
\textrm{BA}&1404&320&36\\\hline
\textrm{TR11}&414&6429&9\\\hline
\textrm{TR41}&878&7454&10\\\hline
\textrm{TR45}&690&8261&10\\\hline
\textrm{TDT2}&9394&36771&30\\\hline
\end{tabular}
\end{table}

%We implement experiments on ten publicly available data sets. The statistics information of these data sets is summarized in Table \ref{data}. Specifically, the first six data sets include four face databases (ORL\footnote{http://www.cl.cam.ac.uk/research/dtg/attarchive/facedatabase.html}, YALE\footnote{http://vision.ucsd.edu/content/yale-face-database}, AR\footnote{http://www2.ece.ohio-state.edu/ aleix/ARdatabase.html}, and JAFFE\footnote{http://www.kasrl.org/jaffe.html}), a toy image database COIL20\footnote{http://www.cs.columbia.edu/CAVE/software/softlib/coil-20.php}, and a binary alpha digits data set BA\footnote{http://www.cs.nyu.edu/~roweis/data.html}. Tr11, Tr41, and Tr45 are derived from NIST TREC Document Database\footnote{http://www-users.cs.umn.edu/{\raise.17ex\hbox{$\scriptstyle\sim$}}han/data/tmdata.tar.gz}. TDT2  corpus\footnote{http://www.cad.zju.edu.cn/home/dengcai/Data/TextData.html} has been among the ideal test sets for document clustering purposes. 
We implement experiments on nine popular data sets. The statistics information of these data sets is summarized in Table \ref{data}. Specifically, the first five data sets include three face databases (ORL, YALE, and JAFFE), a toy image database COIL20, and a binary alpha digits data set BA. Tr11, Tr41, and Tr45 are derived from NIST TREC Document Database. TDT2  corpus has been among the ideal test sets for document clustering purposes. 

Following the setting in \cite{du2015robust}, we design 12 kernels. They are: seven Gaussian kernels of the form $k(x,y)=exp(-\|x-y\|_2^2/(td_{max}^2))$ with $t\in \{0.01, 0.0, 0.1, 1, 10, 50, 100\} $, where $d_{max}$ is the maximal distance between data  points; a linear kernel $k(x,y)=x^\top y$; four polynomial kernels $k(x,y)=(a+x^\top y)^b$ of the form with $a\in\{0,1\}$ and $b\in\{2,4\}$. Besides, all kernels are normalized to $[0,1]$ range, which is done through dividing each element by the largest pairwise squared distance \cite{du2015robust}. 
\subsection{Comparison Methods}
To fully investigate the performance of our method on clustering, we choose a good set of methods to compare. In general, they can be classified into two categories: similarity-based and kernel-based clustering methods.
\begin{itemize}%[noitemsep]
\item{\textbf{Spectral Clustering (SC) }\cite{ng2002spectral}: SC is a widely used clustering technique. It enjoys the advantage of exploring the intrinsic data structures. However, how to construct a good similarity graph is an open issue. Here, we directly use kernel matrix as its input. For our proposed SLSP method, we obtain clustering results by performing spectral clustering with our learned $Z$.}
\item{\textbf{Robust Kernel K-means (RKKM)}\cite{du2015robust}: As an extension to classical k-means clustering method, RKKM has the capability of dealing with nonlinear structures, noise, and outliers in the data. RKKM shows promising results on a number of real-world data sets. }
\item{\textbf{Simplex Sparse Representation (SSR)} \cite{huang2015new}: Based on sparse representation, SSR achieves satisfying performance in numerous data sets. }
%\item{\textbf{Low-Rank Representation (LRR)} \cite{liu2013robust}: Based on self-expression, subspace clustering with low-rank regularizer achieves great success on a number of applications.}
%\item{\textbf{Sparse Subspace Clustering (SSC)} \cite{elhamifar2013sparse}: Based on self-expression, SSC assumes sparse solution of $Z$. Both LRR and SSC learn similarity matrix just by reconstructing the original data, which might lose some important information.}
\item{\textbf{Kernelized LRR (KLRR)} \cite{xiao2016robust}: Based on self-expression, low-rank representation has achieved great success on a number of applications. Kernelized LRR deals with nonlinear data and demonstrates better performance than LRR in many tasks. }
\item{\textbf{Kernelized SSC (KSSC)} \cite{patel2014kernel}: Kernelized version of SSC has also been proposed to capture nonlinear structure information in the input space. Since our framework is an extension of KLRR and KSSC to preserve similarity information, the difference in performance will shed light on the effects of similarity preserving. }
\item{\textbf{Twin Learning for Similarity and Clustering (TLSC)} \cite{kang2017twin}: Based on self-expression, TLSC has been proposed recently and has shown superior performance on a number of real-world data sets. TLSC does not only learn similarity matrix via self-expression in kernel space but also has optimal similarity graph guarantee. However, it fails to preserve similarity information.} %Furthermore, it has good theoretical properties, i.e., it is equivalent to kernel k-means and k-means under certain conditions.
\item{\textbf{SLKE-S} and \textbf{ SLKE-R} \cite{kang2019similarity}: They are closely related to our method developed in this paper. However, they only have similarity preserving term, which might lose some low-oder information.}
\item{Our proposed \textbf{SLSP}: Our proposed structure learning framework with similarity preserving capability. After obtaining similarity matrix $Z$, we perform spectral clustering based on Eq.(\ref{sc}). We examine both low-rank and sparse regularizer and denote their corresponding methods as SLSP-r and SLSP-s, respectively. The implementation of our algorithm is publicly available\footnote{https://github.com/sckangz/L2SP}.}%The code for our method will be publicly available upon acceptance. }
\end{itemize}

\begin{table*}[!ht]
\centering
%\scriptsize
\renewcommand{\arraystretch}{1.6}
\setlength{\tabcolsep}{2.8pt}
 \caption{Clustering results obtained from those benchmark data sets. The average performance of those 12 kernels are put in parenthesis. The best results among those kernels are highlighted in boldface. \label{clusterres}}
\subfloat[Accuracy(\%)\label{acc}]{
\resizebox{.99\textwidth}{!}{
\begin{tabular}{l  |c |  c|c |c| c| c| c| c| c|  c}
	\hline
    Data  & SC &\small{RKKM} &SSR & TLSC &KSSC& KLRR&SLKE-S&SLKE-R   &SLSP-s&SLSP-r  \\
     \hline
        \multirow{1}{*}{YALE}   &49.42(40.52)&48.09(39.71)&54.55&55.85(45.35)&65.45(31.21)&61.21(53.69)&61.82(38.89)&66.24(51.28)&65.45(44.60)&\textbf{66.60}(56.92)\\
	
		\hline
		\multirow{1}{*}{JAFFE}  & 74.88(54.03)&75.61(67.89)&87.32&99.83(86.64)&99.53(35.45)&99.53(90.41)&96.71(70.77)&99.85(90.89)&99.53(82.94)&\textbf{100.0}(93.04)\\
		
		\hline
        \multirow{1}{*}{ORL}  & 58.96(46.65)&54.96(46.88)&69.00&62.35(50.50)&70.50(38.10)&76.50(63.51)&77.00(45.33)&74.75(59.00)&76.50(49.67)&\textbf{81.00}(65.67)\\
%		
%		\hline
%       \multirow{1}{*}{AR}  & 28.82(22.22)&33.43(31.20)&65.00 &56.79(41.35)&41.78(16.03)&77.14(60.10)&&&72.02(45.01)&\textbf{81.19}(64.62)\\
		
		\hline
       \multirow{1}{*}{\small{COIL20}}  & 67.60(43.65)&61.64(51.89)&76.32&72.71(38.03)&73.54(53.54)&83.19(79.48)&75.42(56.83)&84.03(65.65)&\textbf{87.71}(75.94)&\textbf{87.71}(75.58)\\
		
		\hline
        \multirow{1}{*}{BA}& 31.07(26.25)&42.17(34.35)&23.97&47.72(39.50)&50.64(29.29)&47.65(41.07)&50.74(36.35)&44.37(35.79)&\textbf{53.85}(39.51)&52.28(41.96)\\
    
      \hline
                   \multirow{1}{*}{TR11}  & 50.98(43.32)&53.03(45.04)&41.06&71.26(54.79)&62.56(36.94)&79.23(59.60)&69.32(46.87)&74.64(55.07)&68.36(48.53)&\textbf{80.68}(60.23)\\
		
		\hline
		\multirow{1}{*}{TR41}  & 63.52(44.80)&56.76(46.80)&63.78&65.60(43.18)&59.57(33.34)&71.98(58.29)&71.19(47.91)&74.37(53.51)&71.53(53.50)&\textbf{76.80}(60.91)\\
		
		\hline
		\multirow{1}{*}{TR45} & 57.39(45.96)&58.13(45.69)&71.45&74.02(53.38)&71.88(31.87)&78.84(61.18)&78.55(50.59)&79.89(58.37)&79.85(50.08)&\textbf{83.04}(64.18)\\
		
		\hline
		\multirow{1}{*}{TDT2}&52.63(45.26)&48.35(36.67)&20.86&55.74(44.82)&39.82(27.30)&74.80(46.23)&59.61(25.40)&74.92(33.67)&64.98(29.04)&\textbf{75.08}(43.91)\\
\hline
\end{tabular}}
}

%}\\
%%\scriptsize
\renewcommand{\arraystretch}{1.5}
\setlength{\tabcolsep}{2.8pt}
\subfloat[NMI(\%)\label{NMI}]{
\resizebox{.99\textwidth}{!}{
\begin{tabular}{l  |c | c| c| c| c| c| c|  c| c| c}
	\hline
    Data  & SC &\small{RKKM} &SSR & TLSC &KSSC& KLRR &SLKE-S&SLKE-R    &SLSP-s&SLSP-r  \\
       \hline
        \multirow{1}{*}{YALE} & 52.92(44.79)&52.29(42.87)&57.26&56.50(45.07)&63.94(30.03)&62.98(65.91)&59.47(40.38)&64.29(52.87)&\textbf{64.38}(45.36)&64.22(57.07)\\
		
		\hline
		\multirow{1}{*}{JAFFE} & 
82.08(59.35)&83.47(74.01)&92.93&99.35(84.67)&99.17(30.74)&99.16(89.91)&94.80(60.83)&99.49(81.56)&99.17(82.86)&\textbf{100.0}(92.32)\\
		
	\hline
        \multirow{1}{*}{ORL} & 75.16(66.74)&74.23(63.91)&84.23&78.96(63.55)&83.47(28.17)&86.25(78.30)&86.35(58.84)&85.15(75.34)&85.25(61.48)&\textbf{88.21}(79.46)\\
		
%	\hline
%      \multirow{1}{*}{AR} & 58.37(56.05)&65.44(60.81)&84.16&76.02(59.70)&70.40(27.99)&90.38(81.08)&&&87.48(65.86)&\textbf{91.87}(82.39)\\
		
		\hline
       \multirow{1}{*}{COIL20}  &80.98(54.34)&74.63(63.70)&86.89&82.20(73.26)&80.69(59.95)&89.87(78.79)&80.61(65.40)&91.25(73.53)&92.28(83.78)&\textbf{92.36}(83.25)\\
		
	\hline
        \multirow{1}{*}{BA}  &50.76(40.09)&57.82(46.91)&30.29&63.04(52.17)&62.71(54.03)&61.43(55.88)&63.58(55.06)&56.78(50.11)&\textbf{64.76}(57.06)&64.23(55.91)\\
      
	\hline
        \multirow{1}{*}{TR11}  & 43.11(31.39)&49.69(33.48)&27.60&58.60(37.58)&62.92(11.98)&70.82(47.44)&67.63(30.56)&70.93(45.39)&68.06(31.19)&\textbf{72.23}(46.46)\\
		\hline
 \multirow{1}{*}{TR41}  &61.33(36.60)&60.77(40.86)&59.56&65.50(43.18)&63.36(11.57)&69.63(50.26)&\textbf{70.89}(34.82)&68.50(47.45)&68.21(43.43)&70.50(53.02)\\
		
		\hline
		\multirow{1}{*}{TR45}& 48.03(33.22)&57.86(38.96)&67.82&74.24(44.36)&69.23(12.65)&77.01(53.73)&72.50(38.04)&\textbf{78.12}(50.37)&74.26(37.03)&75.27(57.04)\\
		
	\hline
\multirow{1}{*}{TDT2}&52.23(27.16)&54.46(42.19)&02.44&58.35(46.37)&50.65(25.27)&\textbf{73.83}(48.85)&58.55(15.43)&68.21(28.43)&56.10(29.04)&59.77(36.02)\\
		
\hline
\end{tabular}}
} 

\end{table*}
\subsection{Evaluation Metrics}

To quantify the effectiveness of our algorithm on clustering task, we use the popular metrices, i.e., accuracy (Acc) and normalized mutual information (NMI) \cite{cai2009locality}.%, and Purity. 

As the most widely used clustering metric, Acc aims to measure the one-to-one relationship between clusters and classes. If we use $h_i$ and $\hat{h}_i$ to represent the clustering partition and the ground truth label of sample $x_i$, respectively, then we can define Acc as
\[
Acc=\frac{\sum_{i=1}^n \delta(\hat{h}_i, map(h_i))}{n},
\]
where $n$ is the total number of instances, $\delta(\cdot)$ is the famous delta function, and map($\cdot$) maps each cluster index to a true class label  based on Kuhn-Munkres algorithm \cite{chen2001atomic}.
%It is more reliable to measure the imbalanced data sets, i.e., most of objects are from one cluster and only a few objects belong to other clusters. 

The NMI is defined as follows
\[
\textrm{NMI}(H,\hat{H})=\frac{\sum\limits_{h\in H,\hat{h}\in\hat{H}} p(h,\hat{h})\textrm{log}(\frac{p(h,\hat{h})}{p(h)p(\hat{h})})}{\textrm{max}(E(H),E(\hat{H}))},
\]  
where $H$ and $\hat{H}$ denote two sets of clusters, $p(h)$ and $p(\hat{h})$ are the corresponding marginal probability distribution functions induced from the joint distribution $p(h,\hat{h})$, and $E(\cdot)$ represents the entropy function. Bigger NMI value indicates better clustering performance.%NMI takes values between 0 and 1, with 0 and 1 occurring when the two clusters are independent and identical, respectively.

\subsection{Clustering Results}
%To quantitatively assess our algorithm's performance on clustering task, we use the popular measures, i.e., accuracy (Acc) and normalized mutual information (NMI) \cite{cai2009locality}.
We report the experimental results in Table \ref{clusterres}.
As we can see, our method can beat others in almost all experiments. Concretely, we can draw the following conclusions: \\
(i) The improvements of SLSP against SC verify the importance of high quality similarity measure. Rather than directly using kernel matrix in SC, we use learned $Z$ as input of SC. Hence, the big improvement entirely comes from our high-quality similarity measure;\\
 (ii) Comparing SLSP-s with KSSC and SLSP-r with KLRR, we can see the benefit of retaining similarity structure information. In particular, for TDT2 data set, SLSP-s enhances the accuracy of KSSC by 25.16$\%$.\\%In particular, for AR and TDT2 data sets, SLSP-s enhances the accuracy of KSSC by 30.24$\%$ and 25.16$\%$, respectively; \\
(iii) It is worth pointing out our big gain over recently proposed method TLSC. Although both SLSP and TLSC are based on self-expression and kernel method, TLSC fails to consider preserving similarity information, which might be lost during the reconstruction process. \\
 (iv) With respect to SLKE-S and SLKE-R, which have the effect of similarity preserving, our method still outperforms them in most cases. This is attributed to the fact that the first term in Eq. (\ref{obj}) can keep some low-order information, which is missing in SLKE-S and SLKE-R. We can observe that SLSP-r improves the accuracy of SLKE-R over 6\% on ORL, BA, TR11 datasets. \\
In summary, these results confirm the crucial role of similarity measure in clustering and the great benefit due to similarity preserving.
\subsection{Parameter Analysis}
\begin{figure*}[!htbp]
\centering
\includegraphics[width=.48\textwidth]{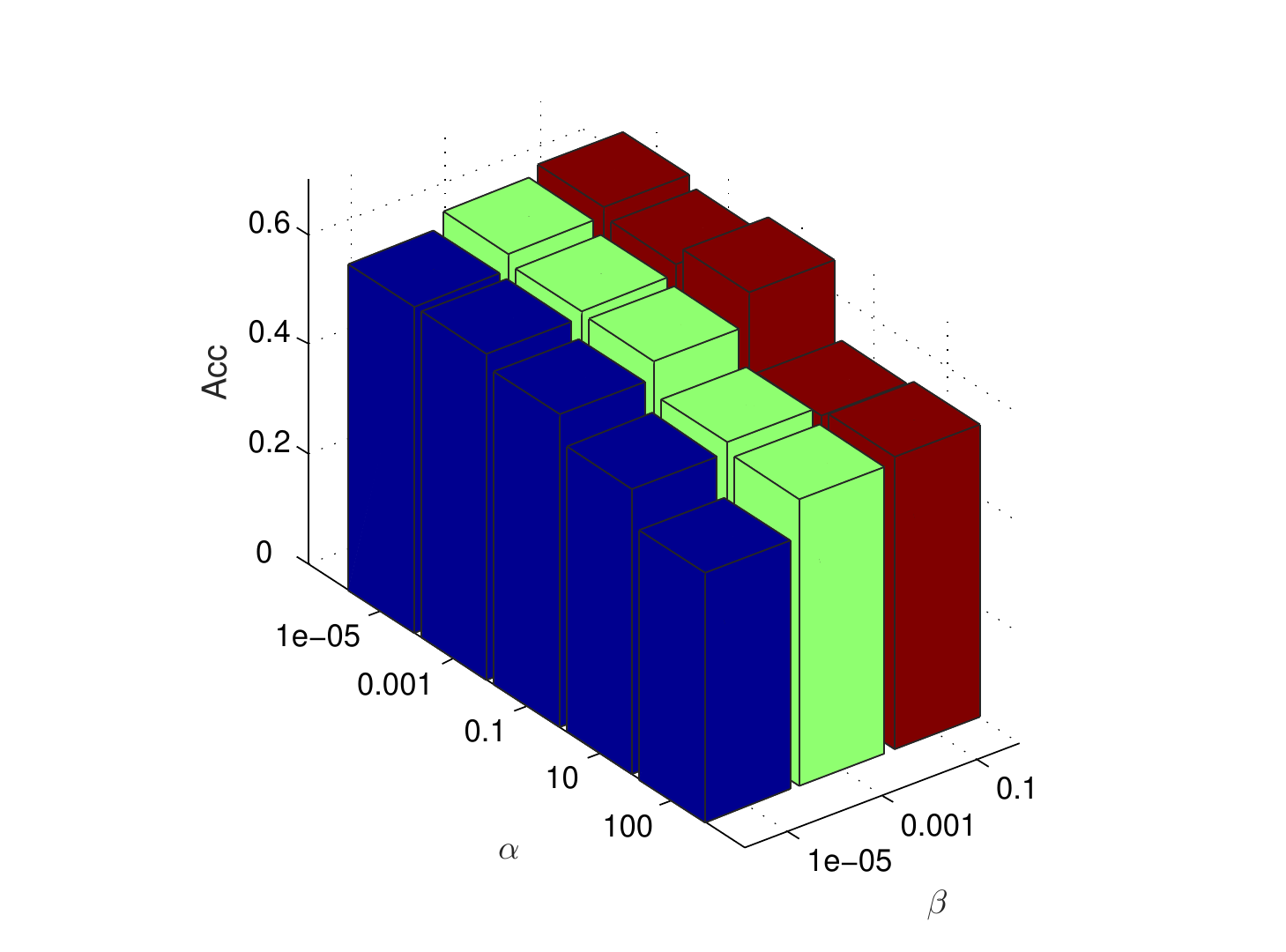}
\includegraphics[width=.48\textwidth]{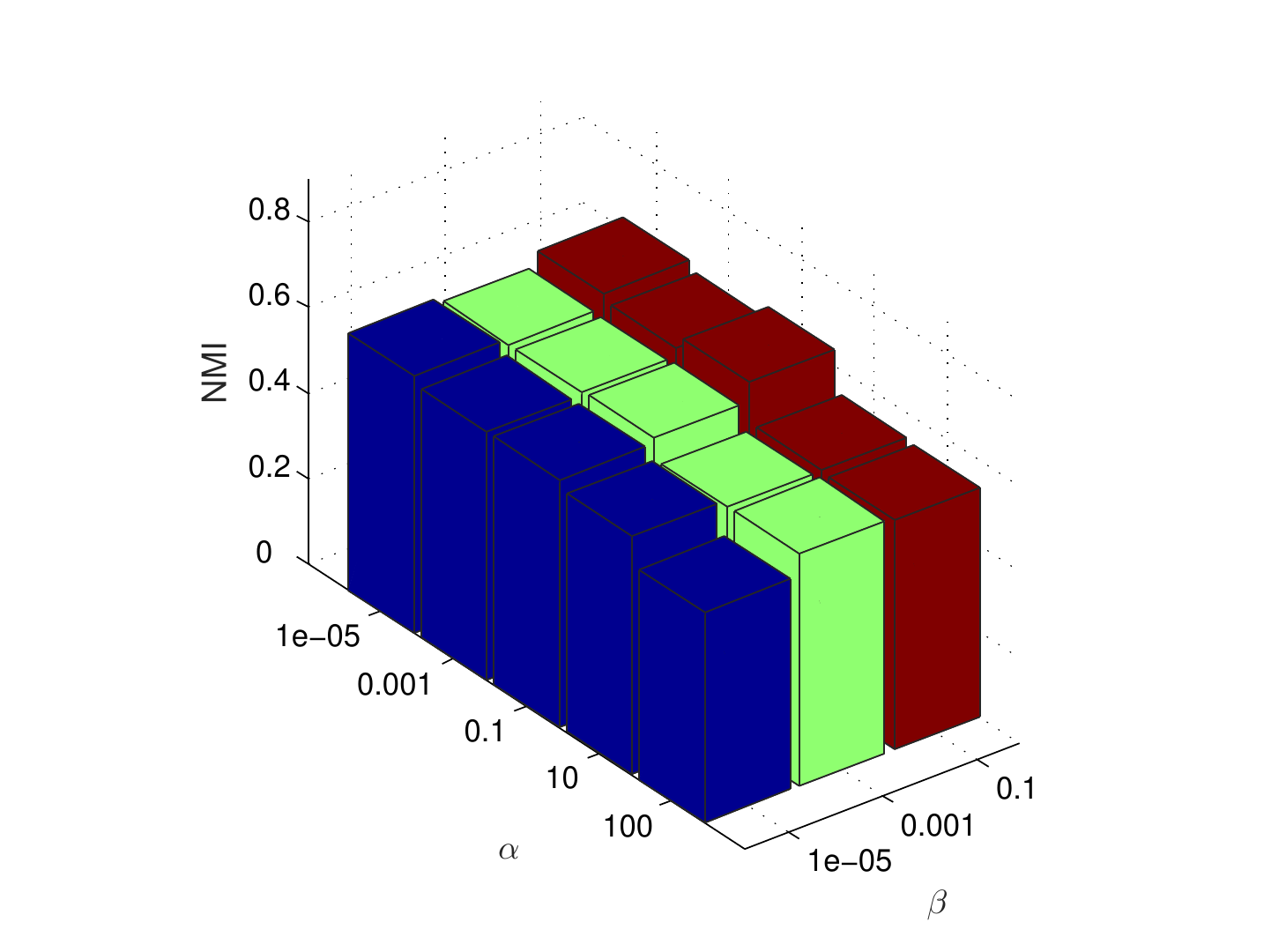}
\caption{Performance of SLSP-r with respect to the variation of $\alpha$ and $\beta$ on YALE data set.}
\label{slker}
\end{figure*}

\begin{figure*}[!htbp]
\centering
\includegraphics[width=.48\textwidth]{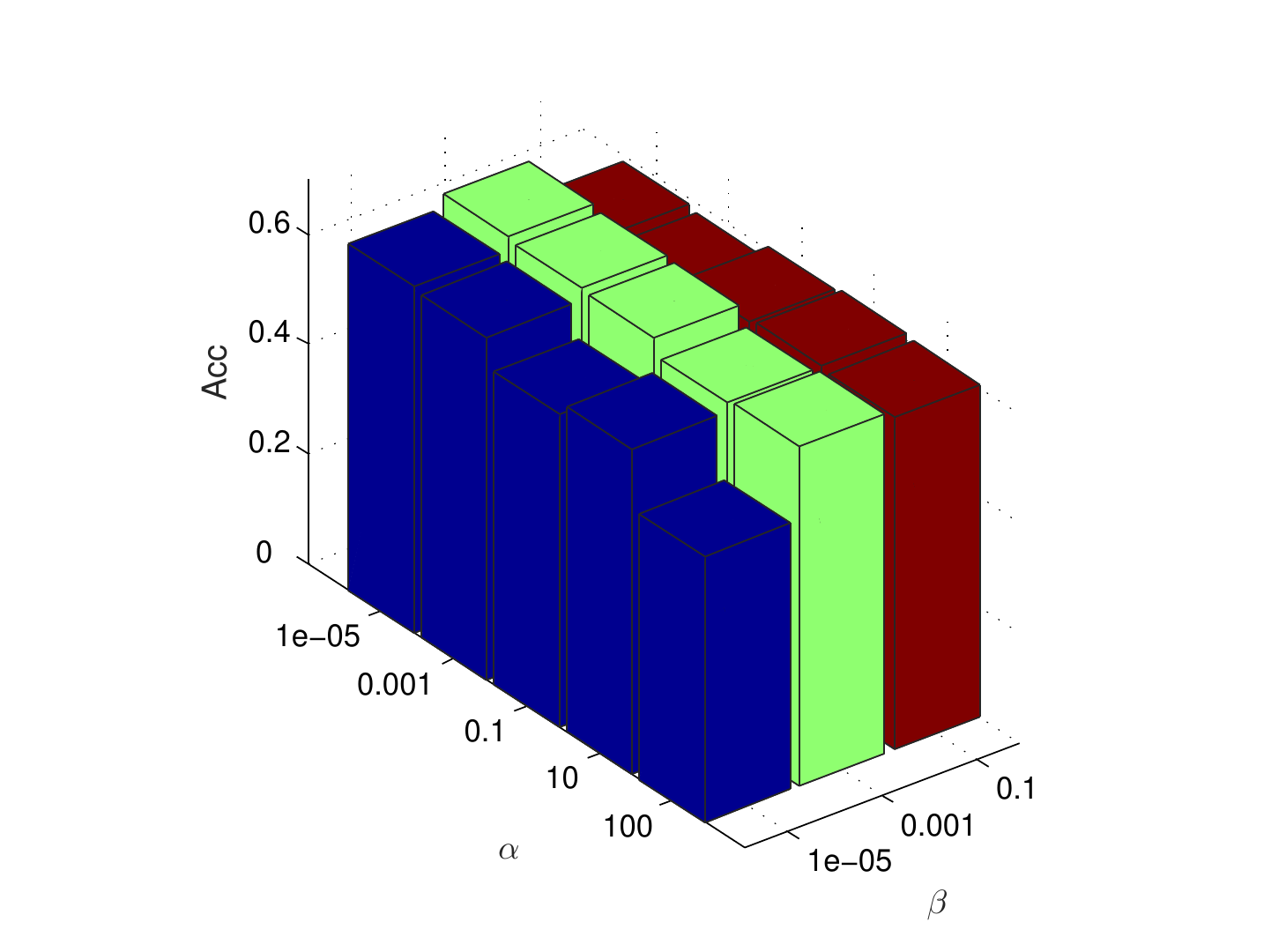}
\includegraphics[width=.48\textwidth]{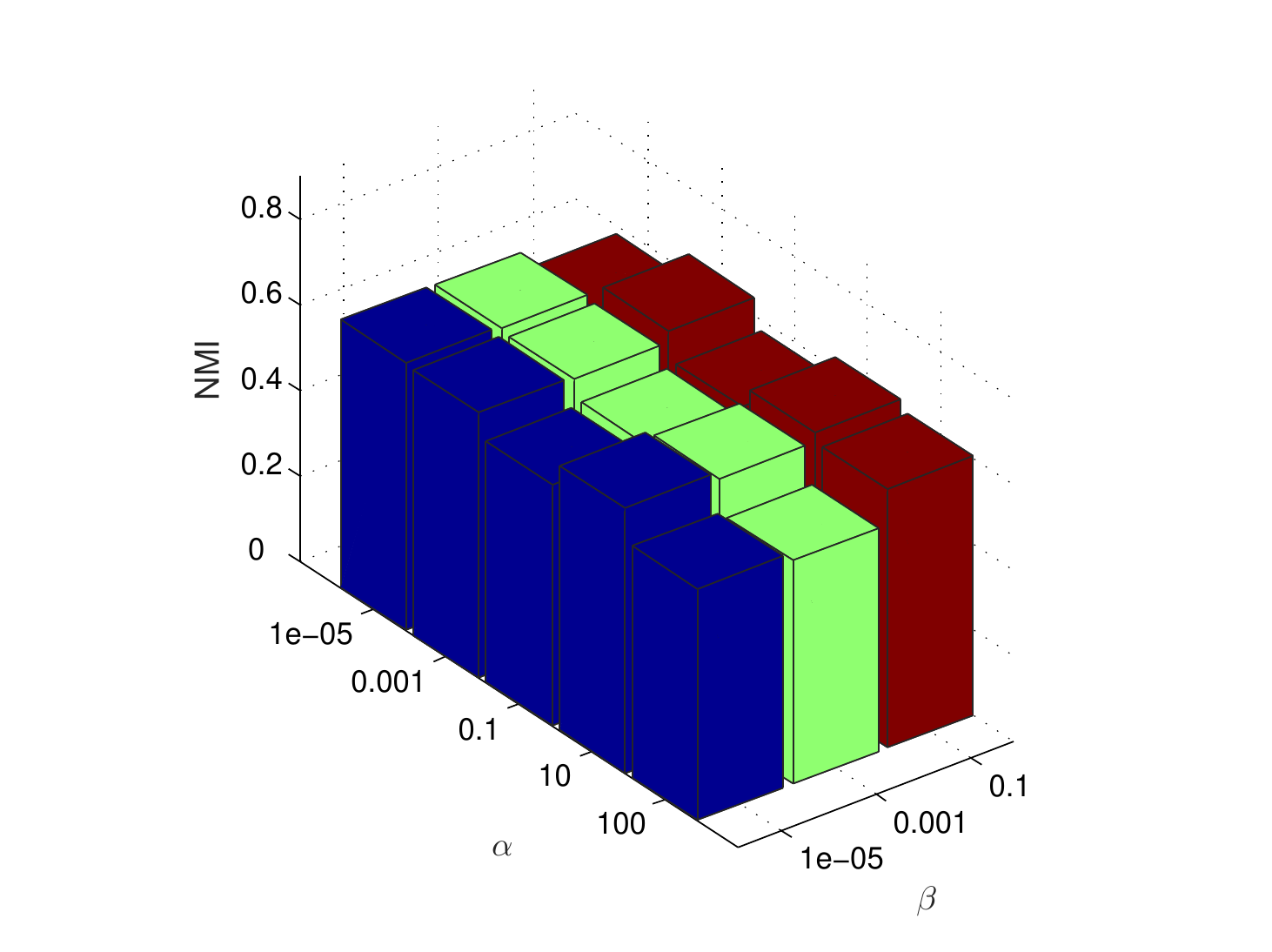}
\caption{Performance of SLSP-s with respect to the variation of $\alpha$ and $\beta$ on YALE data set.}
\label{slkes}
\end{figure*}

There are two parameters in our model: $\alpha$ and $\beta$. Taking YALE data set as an example, we demonstrate the sensitivity of our model SLSP-r and SLSP-s to $\alpha$ and $\beta$ in Figure \ref{slker} and \ref{slkes}. They illustrate that our methods are quite insensitive to $\alpha$ and $\beta$ over wide ranges of values.

%\begin{figure*}[!htbp]
%\centering
%\subfloat[YALE\label{yale}]{\includegraphics[width=.28\textwidth]{yale}}
%\subfloat[BA\label{ba}]{\includegraphics[width=.3\textwidth]{ba}}
%\subfloat[COIL20\label{coil20}]{\includegraphics[width=.34\textwidth]{coil}}
%\caption{Sample images of YALE, BA, and  COIL20.}
%\end{figure*}

\section{Semi-supervised Classification Experiment}
\label{semiexp}
In this section, we show that our method performs well on semi-supervised classification task based on Eq.(\ref{ssgl}).
%\label{class}
%In this section, we show that our methods perform well on semi-supervised classification task.
% In the past decade, graph-based semi-supervised learning (GSSL) has attracted numerous attentions due to its elegant formulation and low computation complexity \cite{cheng2009sparsity}. Similarity graph construction is one of the two fundamental components in GSSL, which is critical to the quality of classification. Nevertheless, with respect to label inference, graph construction has attracted much less attention until recent years \cite{Berton2015}. In this subsection, we demonstrate that our framework can best capture the essential data structure. 
%
%After we obtain graph matrix $Z$ from SLSP, we adopt the popular local and global consistency (LGC) as the classification framework \cite{zhou2004learning}. LGC finds a classification function $F\in \mathcal{R}^{n\times c}$ by solving the following problem:
%\begin{equation}
%\min_F Tr\{F^\top LF+\gamma(F-Y)^\top(F-Y)\},
%\end{equation} 
%where $c$ is the class number, $Y\in \mathcal{R}^{n\times c}$ is the label matrix, in which $y_{ij}=1$ iff the $i$-th sample belongs to the $j$-th class, and $y_{ij}=0$ otherwise. $L$ is the graph Laplacian matrix calculated according to $L=D-Z$, where $D$ is a diagonal matrix with $i$-th element as $\sum_j z_{ij}$.

%Some sample images are shown in Figure \ref{yale}.Figure \ref{ba} shows some sample images from BA database.Some sample images are shown in Figure \ref{coil20}. 
\subsection{Data}
We perform experiments on different types of recognition tasks. \\
(1) \textbf{Evaluation on Face Recognition}: We examine the effectiveness of our similarity graph learning for face recognition on two frequently used face databases: YALE and JAFFE. The YALE face data set contains 15 individuals, and each person has 11 near frontal images taken under different illuminations. Each image is resized to 32$\times$32 pixels. The JAFFE face database consists of 10 individuals, and each subject has 7 different facial expressions (6 basic facial expressions +1 neutral). The images are resized to 26$\times$26 pixels. \\
(2) \textbf{Evaluation on Digit/Letter Recognition}: In this experiment, we address the digit/letter recognition problem on the BA database. The data set consists of digits of ``0" through ``9" and letters of capital ``A" to ``Z", this leads to 36 classes. Moreover, each class has 39 samples. \\
(3) \textbf{Evaluation on Visual Object Recognition}: We conduct visual object recognition experiment on the COIL20 database. The database consists of 20 objects and 72 images for each object. For each object, the images were taken 5 degrees apart as the object is rotating on a turntable.  The size of each image is 32$\times$32 pixels. 

To reduce the work load, we construct 7 kernels for each data set. They include: four Gaussian kernels with $t$ varies over $\{0.1, 1, 10, 100\}$; a linear kernel $k(x,y)=x^\top y$; two polynomial kernels $k(x,y)=(a+x^\top y)^2$ with $a\in\{0,1\}$.
\subsection{Comparison Methods}
We compare our method with several other state-of-the-art algorithms.
\begin{itemize}%[noitemsep]
\item {\textbf{Local and Global Consistency (LGC)} \cite{zhou2004learning}: LGC is a popular label propagation method. For this method, we use kernel matrix as its similarity measure to compute $L$. }
\item{\textbf{Gaussian Field and Harmonic function (GFHF)} \cite{zhu2003semi}: Different from LGC, GFHF is another mechanics to infer those unknown labels as a process of propagating labels through the pairwise similarity.}

\item{\textbf{Semi-supervised Classification with Adaptive Neighbours (SCAN)} \cite{nie2017multi}: Based on adaptive neighbors method, SCAN adds the rank constraint to ensure that $Z$ has exact $c$ connected components. As a result, the similarity matrix and class indicator matrix $F$ are learned simultaneously. It shows much better performance than many other techniques.  }
\item{\textbf{A Unified Optimization Framework for Semi-supervised Learning} \cite{li2015learning}: Li et al. propose a unified framework based on self-expression approach. Similar to SCAN, the similarity matrix and class indicator matrix $F$ are updated alternatively. By using low-rank and sparse regularizer, they have S$^2$LRR and S$^3$R method, respectively.}
\item{\textbf{KLRR} \cite{xiao2016robust} \textbf{and KSSC} \cite{patel2014kernel}: They represent state-of-the-art similarity graph construction techniques. By comparing against them, we can clearly evaluate the effects of similarity preserving on semi-supervised learning.}
\item{Our Proposed\textbf{ SLSP}: After we obtain $Z$ from SLSP-r and SLSP-s, we plug them into LGC algorithm to predict labels for unlabeled data points.}
\end{itemize}

\begin{table*}[htbp]
\begin{center}
\setlength{\tabcolsep}{1pt}
\renewcommand{\arraystretch}{1.5}
\caption{Classification accuracy (\%) on benchmark data sets (mean$\pm$standard deviation). The best results are in bold font.\label{classres}}
\resizebox{.99\textwidth}{!}{
\begin{tabular}{ C{1.15cm}|C{2.1cm}|c|c|c|c|c|c|c|c|c|c}
\hline
Data &Labeled Percentage($\%$) &GFHF & LGC &S$^3$R&S$^2$LRR& SCAN & KLRR&KSSC&SLSP-s& SLSP-r\\
%sets&& &  &&& & &&& \\
\hline\hline
\multirow{3}{4em}{YALE} & 10 &38.00$\pm$11.91&47.33$\pm$13.96& 38.83$\pm$8.60 &28.77$\pm$9.59& 45.07$\pm$1.30 &50.53$\pm$11.36&47.03$\pm$10.32 & 51.20$\pm$1.29& \textbf{53.34}$\pm$14.30\\ 
& 30 & 54.13$\pm$9.47&63.08$\pm$2.20& 58.25$\pm$4.25& 42.58$\pm$5.93& 60.92$\pm$4.03& 62.67$\pm$3.38& 70.08$\pm$3.39 &70.71$\pm$3.13&\textbf{71.13}$\pm$2.88\\ 
& 50 & 60.28$\pm$5.16&69.56$\pm$5.42& 69.00$\pm$6.57& 51.22$\pm$6.78 & 68.94$\pm$4.57& 70.61$\pm$4.98& 77.83$\pm$5.84&\textbf{78.06}$\pm$4.74& 75.89$\pm$4.82\\ 
\hline\hline
\multirow{3}{4em}{JAFFE} & 10 & 92.85$\pm$7.76&96.68$\pm$2.76& 97.33$\pm$1.51& 94.38$\pm$6.23& 96.92$\pm$1.68 & 95.29$\pm$3.27&91.22$\pm$2.46&98.59$\pm$1.07&\textbf{98.99}$\pm$0.83\\ 
& 30 &98.50$\pm$1.01&98.86$\pm$1.14& 99.25$\pm$0.81& 98.82$\pm$1.05& 98.20$\pm$1.22& 99.22$\pm$0.72&98.17$\pm$1.54&\textbf{99.33}$\pm$0.99&99.20$\pm$0.99\\ 
& 50 &98.94$\pm$1.11&99.29$\pm$0.94& 99.82$\pm$0.60& 99.47$\pm$0.59 & 99.25$\pm$5.79& 99.86$\pm$0.32&99.38$\pm$0.65&\textbf{99.91}$\pm$0.27&\textbf{99.91}$\pm$0.99\\ 
\hline\hline
\multirow{3}{4em}{BA} & 10 &45.09$\pm$3.09&48.37$\pm$1.98& 25.32$\pm$1.14 &20.10$\pm$2.51&55.05$\pm$1.67& 46.29$\pm$2.33&49.13$\pm$2.06&56.71$\pm$1.71&\textbf{58.18}$\pm$1.27\\ 
& 30 &62.74$\pm$0.92&63.31$\pm$1.03& 44.16$\pm$1.03& 43.84$\pm$1.54&68.84$\pm$1.09& 62.82$\pm$1.47&66.51$\pm$1.15&\textbf{68.86}$\pm$1.71&67.37$\pm$1.01\\ 
& 50 &68.30$\pm$1.31&68.45$\pm$1.32& 54.10$\pm$1.55& 52.49$\pm$1.27)&72.20$\pm$1.44& 67.74$\pm$1.44&70.69$\pm$1.25&73.40$\pm$1.06&\textbf{73.82}$\pm$1.24\\ 
\hline\hline
\multirow{3}{4em}{COIL20} & 10 &87.74$\pm$2.26&85.43$\pm$1.40& 93.57$\pm$1.59& 81.10$\pm$1.69&90.09$\pm$1.15 & 88.90$\pm$2.46&85.70$\pm$4.03&\textbf{97.35}$\pm$1.22&94.68$\pm$2.38\\ 
& 30 &95.48$\pm$1.40&87.82$\pm$1.03&96.52$\pm$0.68& 87.69$\pm$1.39 &95.27$\pm$0.93& 96.75$\pm$1.49&96.17$\pm$1.65&\textbf{99.46}$\pm$1.55&98.50$\pm$0.85\\ 
& 50 &98.62$\pm$0.71&88.47$\pm$0.45&97.87$\pm$0.10& 90.92$\pm$1.19 &97.53$\pm$0.82& 98.89$\pm$1.02&98.24$\pm$0.97&\textbf{99.91}$\pm$0.34&99.47$\pm$0.59\\ 
\hline
\end{tabular}}
\end{center}

\end{table*}

\subsection{Experimental Setup}
The commonly used evaluation measure accuracy is adopted here. Its definition is
\begin{equation*}
\frac{t}{n}\times 100,
\end{equation*}
where $t$ is the number of samples correctly predicted and $n$ is the total number of samples. We randomly choose some portions of samples as labeled data and repeat 20 times. In our experiment, 10$\%$, 30$\%$, 50$\%$ of samples in each class are randomly selected and labeled. 
Then, classification accuracy and deviation are shown in Table \ref{classres}. For GFHF, LGC, KLRR, KSSC, and our proposed SLSP method, the aforementioned seven kernels are tested and best performance is reported. For these methods, more importantly, the label information is only used in the label propagation stage. For SCAN, S$^2$LRR, and  S$^3$R, the label prediction and similarity learning are conducted in a unified framework, which often leads to better performance.

\subsection{Results}
As expected, the classification accuracy for all methods monotonically increases with the increase of the percentage of labeled samples. As it can be observed, our SLSP method consistently outperforms other state-of-the-art methods. This confirms the effectiveness of our proposed method. Specifically, we have the following observations: \\
(i) By comparing the performance of our proposed SLSP with LGC, we can clearly see the importance of graph construction in semi-supervised learning. On COIL20 data set, the average improvement of SLSP-s and SLSP-r over LGC is 11.67$\%$ and 10.31$\%$, respectively. In our experiments, LGC directly uses kernel matrix as input, while our method uses the learned similarity matrix $Z$ instead in LGC. Hence, the improvements attribute to our high-quality graph construction;\\
 (ii) The superiority of SLSP-s and SLSP-r over KSSC and KLRR, respectively, derives from our consideration of similarity preserving effect. The improvement is considerable especially when the portion of labeled samples is small, which means our method would be promising in a real situation. With 10$\%$ labeling, for example, the average gain is 7.69$\%$ and 6$\%$ for sparse and low-rank representation, respectively;\\
 (iii) Although SCAN, S$^2$LRR, and  S$^3$R can learn similarity matrix and labels simultaneously, our two-step approach still reach higher recognition rate. These imply that our proposed method can produce a more accurate similarity graph than existing techniques that without explicit similarity preserving capability.
\section{Deep Clustering Experiment}
\label{deepexp}
To demonstrate the effect of deep model DSLSP, we follow the settings in \cite{ji2017deep} and perform clustering task on Extended Yale B (EYaleB), ORL, COIL20, and COIL40 datasets. We compare with LRR \cite{liu2013robust}, Low Rank Subspace Clustering (LRSC) \cite{vidal2014low}, SSC \cite{elhamifar2013sparse}, Kernel Sparse Subspace Clustering (KSSC) \cite{patel2014kernel}, SSC by Orthogonal Matching Pursuit (SSC-OMP) \cite{you2016scalable}, Efficient Dense Subspace Clustering (EDSC) \cite{ji2014efficient}, SSC with pre-trained convolutional auto-encoder features (AE+SSC), Deep Embedding Clustering (DEC) \cite{xie2016unsupervised}, Deep $k$-means (KDM) \cite{fard2018deep}, Deep Subspace Clustering Network with $\ell_1$ norm (DSC-Net-L1) \cite{ji2017deep}, and Deep Subspace Clustering Network with $\ell_2$ norm (DSC-Net-L2) \cite{ji2017deep}. For a fair comparison with DSC-Nets, we adopt $\ell_1$ and $\ell_2$ norm respectively using the same network architectures, which are denoted as DSLSP-L1 and DSLSP-L2. We adopt convolutional neural networks (CNNs) to implement the auto-encoder. Adam is employed to do the optimization \cite{kingma2014adam}. The full batch of dataset is fed to our network. We pre-train the network without the self-expression layer. The details of the network structures are shown in Table \ref{netstruc}.

\begin{table*}[htbp]
\begin{center}
\caption{Network settings for our experiments, including the "kernel size@channels" and size of $Z$.\label{netstruc}}
\resizebox{.8\textwidth}{!}{
\begin{tabular}{ C{1.15cm}|C{2.1cm}|c|c|c}
\hline
 &EYaleB&ORL & COIL20& COIL40 \\
\hline %\hline

\multirow{3}{4em}{encoder}&5$\times$5@10&5$\times$5@5&3$\times$3@15&3$\times$3@20\\
&3$\times $3@20&3$\times$3@3&-&-\\
&3$\times $3@30&3$\times $3@3&-&-\\
\hline
Z&2432$\times$2432 &400$\times$400&1440$\times$1440&2880$\times$2880\\
\hline
\multirow{3}{4em}{decoder}&3$\times $3@30&3$\times $3@3&3$\times$3@15&3$\times$3@20\\
&3$\times $3@20&3$\times $3@3&-&-\\
&5$\times $5@10&5$\times $5@5&-&-\\
\hline
\end{tabular}}
\end{center}
\end{table*}

\begin{table*}[htbp]
\centering
\caption{Clustering results on EYaleB, ORL, COIL20, and COIL40 \label{deepres}}
\renewcommand{\arraystretch}{1.5}
\resizebox{.99\textwidth}{!}{
\begin{tabular}{c |c|c  c c c c c|c c c c c c c}
\hline
Dataset & Metric & SSC  & KSSC & SSC-OMP & EDSC & LRR & LRSC & AE+SSC&DEC&DKM & DSC-Net-L1 & DSC-Net-L2 & DSLSP-L1 & DSLSP-L2 \\
\hline
\multirow{2}{4em}{EYaleB} & Accuracy & 0.7354  & 0.6921 & 0.7372 & 0.8814 & 0.8499 & 0.7931 & 0.7480&0.2303&0.1713 & 0.9681 & 0.9733 & 0.9757& \textbf{0.9762}\\
 & NMI & 0.7796 & 0.7359 & 0.7803 & 0.8835 & 0.8636 & 0.8264 & 0.7833&0.4258&0.2704 & 0.9687 & \textbf{0.9703} & 0.9668 & 0.9674\\
%& PUR & 0.7467  & 0.7183 & 0.7542 & 0.8800 & 0.8623 & 0.8013 & 0.7597 & 0.9711 & 0.9731 & 0.9766 & 0.9762\\
\hline

\multirow{2}{4em}{ORL}& Accuracy & 0.7425  & 0.7143 & 0.7100 & 0.7038 & 0.8100 & 0.7200 & 0.7563 &0.5175&0.4682& 0.8550 & 0.8600 & 0.8700 & \textbf{0.8775}\\
 & NMI & 0.8459  & 0.8070 & 0.7952 & 0.7799 & 0.8603 & 0.8156 & 0.8555&0.7449&0.7332 & 0.9023 & 0.9034 & 0.9237 & \textbf{0.9249}\\
%& PUR & 0.7875  & 0.7513 & 0.7463 & 0.7138 & 0.8225 & 0.7542 & 0.7950 & 0.8585 & 0.8625 & 0.8825 & 0.8900\\
\hline

\multirow{2}{4em}{COIL20 }& Accuracy & 0.8631  & 0.7087 & 0.6410 & 0.8371 & 0.8118 & 0.7416 & 0.8711&0.7215&0.6651 & 0.9314 & 0.9368 & 0.9743 & \textbf{0.9757}\\
& NMI & 0.8892 & 0.8243 & 0.7412 & 0.8828 & 0.8747 & 0.8452 & 0.8990 &0.8007&0.7971& 0.9353 & 0.9408 & 0.9731 & \textbf{0.9740}\\
%& PUR & 0.8747  & 0.7497 & 0.6667 & 0.8585 & 0.8361 & 0.7937 & 0.8901 & 0.9306 & 0.9397 & 0.9757 & 0.9757\\
\hline
\multirow{2}{4em}{COIL40 }& Accuracy & 0.7191  & 0.6549 & 0.4431 & 0.6870 & 0.6493 & 0.6327 & 0.4872&0.5812&0.1713 & 0.8003 & 0.8075 & 0.8389 & \textbf{0.8417}\\
& NMI & 0.8212 & 0.7888 & 0.6545 & 0.8139 & 0.7828 & 0.7737 & 0.8318&0.7417&0.7840 & 0.8852 & 0.8941 & 0.9262 & \textbf{0.9267}\\
 \hline
\end{tabular}}
\end{table*}

The clustering performance of different methods is provided in Table \ref{deepres}. We observe that DSLSP-L2 and DSLSP-L1 achieve very good performance. Specifically, we have the following observations:
\begin{itemize}
\item The $\ell_2$ norm performs slightly better than $\ell_1$ norm. This is consistent with the results in \cite{ji2017deep}. Perhaps, this is caused by the inaccurate optimization in $\ell_1$ norm since it is non-differentiable at zero.
\item As they share the same network for latent representation learning, the improvement of DSLSP over DSC-Net is attributed to our introduced similarity preserving mechanism. Note that the only difference between their objective function is the additional similarity preserving term in Eq. (\ref{deepobj}). For example, on COIL20, DSLSP-L2 improves over DSC-Net-L2 by 3.89\% and 3.32\% in terms of accuracy and NMI, respectively. For COIL40, our method with $\ell_2$ norm outperforms DSC-Net-L2 by 3.42\% on accuracy and 3.26\% on NMI.
\item Both ORL and COIL20 datasets are used in Table \ref{clusterres} and \ref{deepres}. DSLSP-L2 enhances the accuracy from 0.81, 0.8771 in Table \ref{clusterres} to 0.8775 and 0.9757, respectively. Once again, this demonstrates the power of deep learning models. Furthermore, for these two datasets, our results in Table \ref{clusterres} are also better than the shallow methods and AE+SSC in Table \ref{deepres}. This further verifies the superior advantages of our similarity preserving approach.
\item Compared to DEC and DKM, our method can improve the performance significantly. This is own to that our method is based on similarity, while other methods are based on Euclidean distance which is not suitable for complex data.
\end{itemize}
In summary, above conclusions imply the superiority of our proposed similarity preserving term, no matter in shallow or deep models. 
\section{Conclusion}
\label{conclusion}
In this paper, we introduce a new structure learning framework, which is capable of obtaining highly informative similarity graph for clustering and semi-supervised methods. Different from existing low-dimensional structure learning techniques, a novel term is designed to take advantage of sample pairwise similarity information in the learning stage. In particular, by incorporating the similarity preserving term in our objective function, which tends to keep the similarities between samples, our method consistently and significantly improves clustering and classification accuracy. Therefore, we can conclude that our framework can better capture the geometric structure of the data, resulting in more informative and discriminative similarity graph. Besides, our method can be easily extended to other self-expression based methods. In the future, we plan to further investigate efficient algorithms for constructing large-scale similarity graphs. Also, current methods conduct label learning after graph construction. It is interesting to develop principled method to solve the graph construction and label learning problems at the same time.
\section*{Acknowledgment}
This paper was in part supported by Grants from the Natural
Science Foundation of China (Nos. 61806045 and 61572111) and a Fundamental
Research Fund for the Central Universities of China (Nos. ZYGX2017KYQD177).

%
%\ifCLASSOPTIONcaptionsoff
%  \newpage
%\fi

%received his M.S. degree in physics from Sichuan University, China, in 2011. He

\bibliographystyle{elsarticle-num}
\bibliography{cvprref}

\end{document}